%% file: main.tex
\documentclass[sigconf]{acmart}

\input{preamble/00_packages.tex}

\input{preamble/01_commands.tex}

\setcopyright{acmlicensed}
\copyrightyear{2027}
\acmYear{2027}
\acmDOI{XXXXXXX.XXXXXXX}
\acmConference[KDD '27]{The 33rd ACM SIGKDD Conference on Knowledge Discovery and Data Mining}{August 1--5, 2027}{San Jose, USA}
\acmISBN{978-1-4503-XXXX-X/2018/06}

\begin{document}

\input{sections/0.1_title.tex}
\input{sections/0.2_author.tex}
\input{sections/0.3_abs.tex}
\input{sections/0.4_ccsxml.tex}
\input{sections/0.5_keywords.tex}

\maketitle

\input{sections/1_intro.tex}
\input{sections/2_pre.tex}
\input{sections/3_method.tex}
\input{sections/4_exp.tex}
\input{sections/5_deployment.tex}
\input{sections/6_related.tex}
\input{sections/7_conclusion.tex}

\bibliographystyle{ACM-Reference-Format}
\bibliography{ref}

\end{document}

%% file: preamble/00_packages.tex
\usepackage{enumitem} 
\usepackage{subcaption} 
\usepackage{xspace} 
\usepackage{threeparttable} 

\usepackage{multirow}
\usepackage{makecell}

\usepackage{algorithm}
\usepackage{algpseudocode}

\usepackage[most]{tcolorbox}
\usepackage{listings}

%% file: preamble/01_commands.tex
\newcommand{\paratitle}[1]{\vspace{1ex}\noindent{\bf #1}}

\definecolor{deploymentrevision}{RGB}{128,0,128}

\definecolor{constructionFrame}{HTML}{9A7148}
\definecolor{constructionBack}{HTML}{FDF8F0}
\definecolor{analysisFrame}{HTML}{597A96}
\definecolor{analysisBack}{HTML}{F3F7FA}
\definecolor{preparationFrame}{HTML}{7B6D91}
\definecolor{preparationBack}{HTML}{F7F4FA}
\definecolor{updateFrame}{HTML}{66856A}
\definecolor{updateBack}{HTML}{F3F8F3}

\newtcblisting{promptbox}[3]{
  enhanced,
  breakable,
  listing only,
  colframe=#1,
  colback=#2,
  coltitle=white,
  boxrule=0.8pt,
  arc=1.5mm,
  left=1.5mm,
  right=1.5mm,
  top=1mm,
  bottom=1mm,
  fonttitle=\normalsize\bfseries,
  title={#3},
  width=\linewidth,
  listing options={
    basicstyle=\scriptsize\ttfamily,
    breaklines=true,
    columns=fullflexible,
    keepspaces=true,
    breakatwhitespace=false,
    showstringspaces=false
  }
}

\usepackage{xcolor}

%% file: sections/0.1_title.tex
\title[AgentOnto: Ontology-Guided LLM Agents for Enterprise Data Analytics]{AgentOnto: Ontology-Guided LLM Agents \\ for Enterprise Data Analytics}

%% file: sections/0.2_author.tex
\author{
Dayan Pan$^{1}$,
Bincheng Zuo$^{2}$,
Xiaojian Wang$^{2}$,
Daogang Zhang$^{2}$,
Jingyuan Wang$^{1}$
}

\affiliation{
\institution{
$^{1}$School of Computer Science and Engineering, Beihang University, Beijing, China\\
$^{2}$Omaha AIGC, Shanghai, China
}
\country{}
}

\renewcommand{\authors}{Dayan Pan, Bincheng Zuo, Xiaojian Wang, Daogang Zhang, Jingyuan Wang}

\renewcommand{\shortauthors}{Dayan Pan, Bincheng Zuo, Xiaojian Wang, Daogang Zhang, Jingyuan Wang}

%% file: sections/0.3_abs.tex
\begin{abstract}
Enterprise data analysis requires agents to connect analytical requests with knowledge distributed across databases, business documents, and existing example tasks. Without an explicit representation of this knowledge, an agent may produce incorrect
analytical results and cannot reliably incorporate feedback.
We present AgentOnto, an ontology-guided LLM agent for enterprise data analytics.
AgentOnto organizes heterogeneous enterprise knowledge into an ontology that directly guides analysis and incorporates confirmed feedback. Iterative Ontology Preparation further improves the ontology through self-generated analysis tasks. 
We evaluate AgentOnto on two public data sources. AgentOnto substantially outperforms representative methods, while module-level evaluations confirm the effectiveness of ontology construction, analysis, and updating.
A real-world deployment further demonstrates its practical value in enterprise analytics. 
These results demonstrate that explicitly constructing, using, and updating enterprise knowledge supports reliable and adaptable data analysis.
\end{abstract}

%% file: sections/0.4_ccsxml.tex
\ccsdesc[500]{Information systems~Data mining}
\ccsdesc[500]{Applied computing~Enterprise ontologies, taxonomies and vocabularies}

%% file: sections/0.5_keywords.tex
\keywords{Enterprise Data Analytics, Large Language Model Agents, Ontology-Based Analysis}

%% file: sections/1_intro.tex
\input{figures/intro_analysis.tex}
\section{Introduction} \label{sec:intro}

Enterprise decision-making increasingly depends on data analysis. As enterprise data sources, business knowledge, and analytical requirements grow, predefined dashboards and fixed workflows can no longer meet the breadth and complexity of analytical needs. Large language model (LLM) agents offer a flexible analytical interface by combining language understanding, reasoning, database access, and tool use. Recent enterprise business-intelligence systems, including SiriusBI~\cite{jiang2024siriusbi} and CoeusBI~\cite{lian2026coeusbi}, demonstrate the practical value of this agent-based approach for enterprise analytics.

However, accurate enterprise analytics requires an agent to identify the relevant business entities and relationships and apply them correctly to the available data.
Ontologies provide explicit conceptual models that organize this business knowledge and connect it to enterprise databases.
Ontology-based data access systems such as SODA~\cite{blunschi2012soda} and ATHENA~\cite{saha2016athena,sen2020athena++} use these models and mappings to translate user questions into database operations~\cite{xiao2018ontology}.
In parallel, LLM data-analysis agents extend natural-language querying with planning, tool invocation, and execution feedback, and recent work uses LLMs to construct ontologies or reusable knowledge~\cite{giglou2023llms4ol,rissaki2024towards}. 
These studies connect ontologies to enterprise data, enabling broader analytical capabilities for agents.
Despite these advances, building ontology-guided agents for enterprise analytics still presents three key challenges.

\textbf{(1) Heterogeneous Sources Complicate Ontology Construction.} Enterprise analytics knowledge is distributed across databases, business documents, and existing example tasks, which respectively describe data structure, business concepts, and analysis patterns. 
A well-constructed enterprise analytics ontology integrates knowledge from these sources around entities and relationships with supporting evidence.
Existing methods, however, typically focus on individual sources~\cite{oyewale2026llm} or ontology components~\cite{lippolis2025ontology}, providing limited support for integrating heterogeneous evidence.

\textbf{(2) Insufficient Ontology Integration Constrains Agent Analytics.} 
Providing an ontology to an agent does not ensure correct analytical behavior.
Existing approaches use ontologies to support query generation or checking~\cite{sen2020athena++,allemang2024increasing}, but high answer coverage does not guarantee correctness because an agent may return incorrect results or answer requests that should not be executed, leading to silent semantic errors.
Effective ontology guidance should therefore support both correct analytical decisions and accurate execution results.

\textbf{(3) Static Ontologies Limit Continuous Adaptation.} Enterprise knowledge evolves over time, and analytical feedback can reveal missing or incorrect ontology definitions. Existing evolution and editing methods help revise ontologies~\cite{noy2004ontology,flouris2008ontology,zablith2009evolva,tudorache2013webprotege,qiang2024om4ov}, but provide limited support for validating their effects on subsequent analyses. Effective ontology updating should therefore correct the targeted problem while preserving analytical behavior that should remain unchanged.

Our analysis of two public retail and insurance workloads in Figure~\ref{fig:analysis} provides further empirical evidence for these challenges~\cite{rumiantsau2026semantic,sequeda2024benchmark}. Their tasks combine multiple analytical operations, and the retail workload requires 46 business concepts not captured by its database schema. These findings motivate AgentOnto to organize enterprise knowledge as an explicit ontology that guides analysis and is refined through analytical feedback.

To address these challenges, we present \textbf{AgentOnto}, an ontology-guided LLM agent for enterprise data analytics.
AgentOnto integrates ontology construction, data analysis, and ontology updating into a unified workflow, with an enterprise analytics ontology guiding the entire process.
(i)\emph{Ontology Construction} integrates knowledge from enterprise data, business documents, and existing example tasks into an enterprise analytics ontology with supporting evidence. 
(ii)\emph{Ontology-Guided Analysis} uses the ontology to guide plan formation, checking, and execution.
(iii)\emph{Ontology Updating} turns confirmed feedback into targeted changes and verifies that the target problem is corrected while other established behavior is preserved.
Building on these methods, AgentOnto introduces \emph{Iterative Ontology Preparation}, which uses self-generated analysis tasks to examine and refine the ontology before handling actual analysis tasks.

We evaluate AgentOnto using public data from two sources: the World Development Indicators (WDI) and the U.S. EPA Air Quality System (AQS). We compare it with methods based on static ontologies, multi-stage SQL generation, and database-agent tool use. AgentOnto achieves an analysis accuracy of 87.5\%, compared with 55.0\% for the strongest baseline. Evaluations of the three core methods further show that AgentOnto improves ontology construction, analysis, and updating across both domains.
Taken together, these results show that an explicit enterprise analytics ontology can improve reliability across related requests by guiding analysis and incorporating changes in business requirements.
We further quantify AgentOnto through a real-world deployment, demonstrating its ability to support reliable and adaptable enterprise data analysis.

The main contributions of this work are as follows:

\begin{itemize}[leftmargin=*, topsep=0pt]
\item We establish ontology-guided LLM agents as a systematic approach to enterprise data analytics, using an explicit and maintainable representation of enterprise knowledge to support reliable analysis.

\item We propose AgentOnto, which organizes heterogeneous enterprise knowledge into an ontology that directly guides analysis and incorporates feedback. Iterative Ontology Preparation further improves the ontology through self-generated tasks.

\item 
We conduct extensive experiments on public datasets, showing that AgentOnto improves overall analysis reliability and the effectiveness of its core methods over representative approaches.
A real-world deployment further demonstrates its practical value in enterprise analytics.
\end{itemize}

%% file: figures/intro_analysis.tex
\begin{figure}[ht]
    \centering
    \includegraphics[width=1.\linewidth]{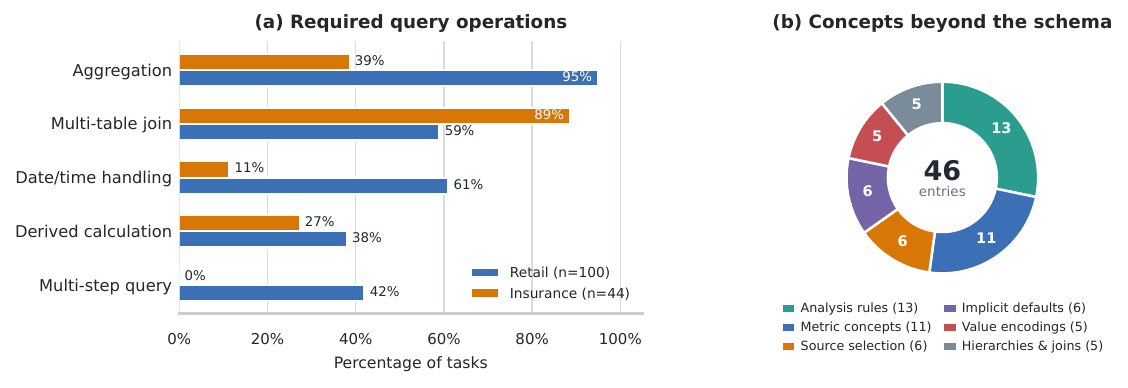}
    \caption{Analytical operations and business concepts in two public workloads. (a) Share of tasks requiring different analytical operations across the retail and insurance workloads. (b) Distribution of business concepts required by the retail workload but not captured by its database schema.}
    \label{fig:analysis}
\end{figure}

%% file: sections/2_pre.tex
\input{figures/model_arch.tex}

\section{Preliminaries} \label{sec:pre}

\subsection{Enterprise Analytics Setting}
Enterprise data analysis answers a given request by interpreting its requirements, accessing and processing the relevant enterprise data, and producing an analytical result.
Three enterprise resources support this process. The \emph{enterprise database} ($\mathsf{DB}$) contains the database schema and data records available for analysis. \emph{Business documents} ($\mathsf{Docs}$) describe business concepts, metrics, units, applicability conditions, and analytical rules. \emph{Existing example tasks} ($\mathsf{Examples}$) show how this knowledge is applied to data in practice.

An analytical request \(\mathsf{req}_i\) expresses a user's analytical objective in natural language. The request may not contain all the information needed for analysis, so processing it requires identifying the relevant enterprise knowledge and connecting that knowledge to \(\mathsf{DB}\). 
A successfully executed query may still produce a \emph{silent semantic error} when its result has an incorrect business meaning.

\subsection{Ontologies in Enterprise Analytics}

An ontology explicitly represents the concepts in a domain, the relationships among them, and the rules for their use.
In enterprise analytics, an ontology organizes reusable business knowledge and connects it to enterprise databases. It thereby provides a common representation for interpreting different analytical requests over the same data.

We represent the enterprise analytics ontology at state $t$ as
\begin{equation}
O_t = (E_t, M_t, G_t),
\label{eq:ontology}
\end{equation}
where the three components capture complementary forms of enterprise knowledge:

\begin{itemize}[leftmargin=*, topsep=0pt]
\item $E_t$, \textbf{Entities and Relationships}, represents business entities, their attributes and values, and the relationships among them.

\item $M_t$, \textbf{Metric Definitions}, represents the business meaning and calculation of analytical metrics, together with the conditions for their use.

\item $G_t$, \textbf{Database Mappings}, connects entities, attributes, relationships, and metrics to the tables, columns, and join paths in $\mathsf{DB}$.
\end{itemize}

Each ontology entry \(x\in E_t\cup M_t\cup G_t\) records its supporting sources \(\mathcal{S}(x)\) and confirmation status \(c(x)\). The sources identify the evidence and its location, while \(c(x)\) indicates whether the entry has been confirmed based on that evidence.

\subsection{Ontology-Guided Data Analysis}

Given an analytical request $\mathsf{req}_i$, an ontology-guided agent uses $O_t$ to connect the request with the enterprise data in $\mathsf{DB}$. Entities, relationships, and metric definitions provide the business knowledge needed to interpret the request, while database mappings connect that interpretation to executable data operations. The agent can then form an analysis plan that combines reusable ontology knowledge with the choices specific to the request.

The ontology guides both the interpretation of the request and its execution over enterprise data. It also helps the agent recognize when the available information is insufficient for producing a valid result. Ontology-guided data analysis therefore supports consistent analytical decisions and results across requests that rely on the same enterprise knowledge.

%% file: figures/model_arch.tex
\begin{figure*}[t]
    \centering
    \includegraphics[width=0.9\textwidth]{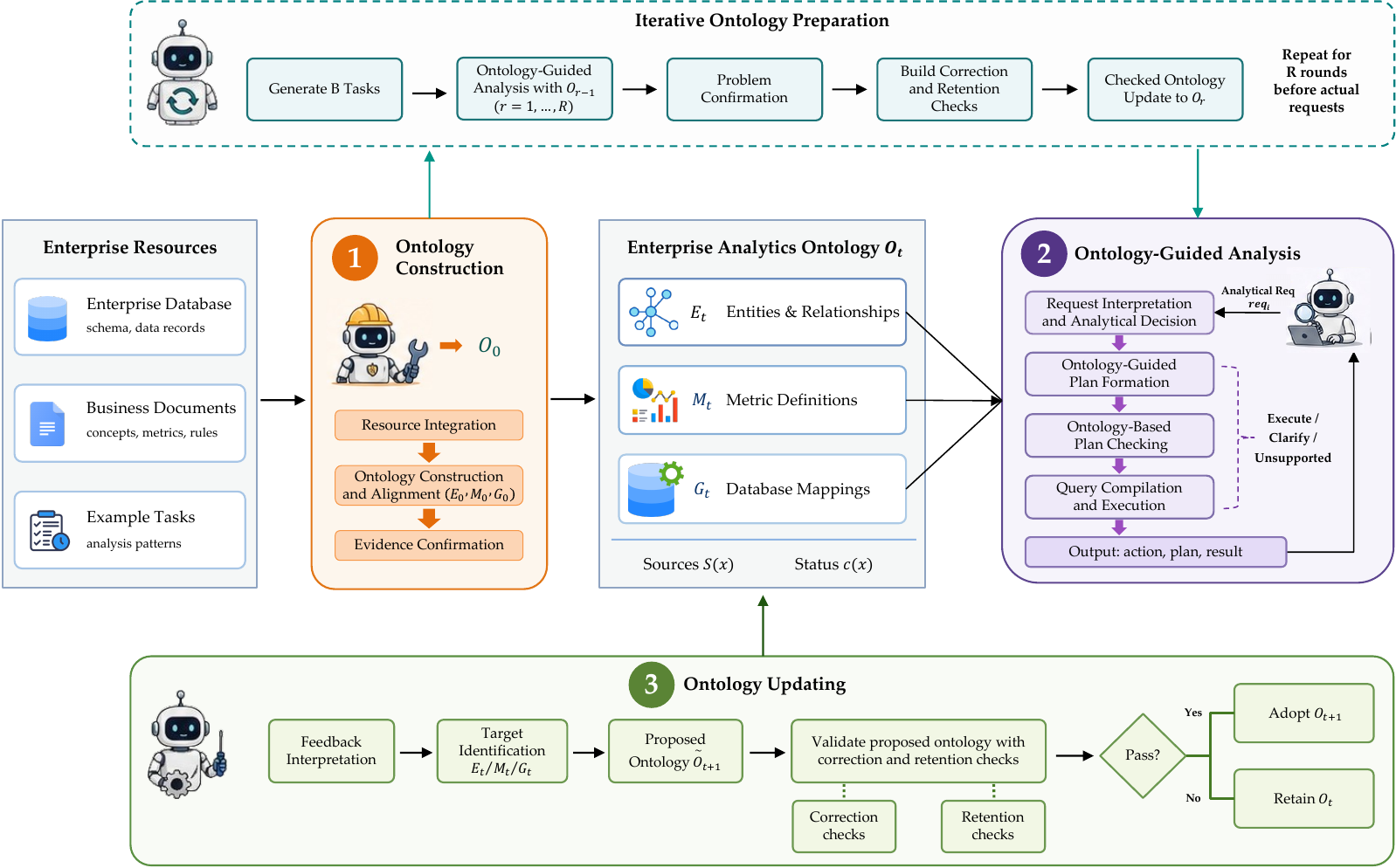}
    \caption{Overview of AgentOnto. Ontology Construction integrates enterprise resources into an enterprise analytics ontology. Ontology-Guided Analysis uses the ontology throughout request processing, while Ontology Updating incorporates confirmed feedback through targeted changes and behavioral checks. Iterative Ontology Preparation repeatedly generates and analyzes tasks to examine and refine the ontology before actual requests.}
    \label{fig:framework}
\end{figure*}

%% file: sections/3_method.tex
\section{The AgentOnto Framework} \label{sec:method}

\subsection{Overview}

Enterprise analytics agents need to integrate knowledge distributed across enterprise data, business documents, and existing example tasks, apply this knowledge throughout analysis, and incorporate changes revealed by analytical feedback. AgentOnto addresses these requirements by organizing enterprise knowledge as the enterprise analytics ontology defined in Equation~\ref{eq:ontology} and using the ontology to guide the entire analytical process.

Figure~\ref{fig:framework} presents the AgentOnto framework, which consists of three core methods. \emph{Ontology Construction} integrates the enterprise resources into an ontology containing entities and relationships, metric definitions, and database mappings, with supporting evidence for each entry. Given an analytical request, \emph{Ontology-Guided Analysis} uses the ontology to form and check an analysis plan, connect the plan to the database, and produce the analytical result. When confirmed feedback reveals an ontology problem, \emph{Ontology Updating} makes a targeted change and verifies both the intended correction and the analytical behavior that should remain unchanged.
Building on these methods, \emph{Iterative Ontology Preparation} uses self-generated analysis tasks to examine and refine the constructed ontology before actual requests are handled.

Together, these methods allow heterogeneous enterprise knowledge to be organized as an ontology, directly used during analysis, and improved through analytical feedback. Iterative Ontology Preparation further examines the ontology through analytical use before it supports actual requests. The following subsections describe these methods and the complete AgentOnto workflow.

\subsection{Ontology Construction, Analysis, and Updating}

\paratitle{Ontology Construction.}
Enterprise analytics knowledge is distributed across resources that describe different aspects of analysis. The enterprise database \(\mathsf{DB}\) contains the database schema and data records, business documents \(\mathsf{Docs}\) describe business concepts, metrics, and analytical rules, and existing example tasks \(\mathsf{Examples}\) show how this knowledge is applied in analysis. Ontology Construction integrates these complementary resources to produce the initial enterprise analytics ontology:
\begin{equation}
O_0 = \operatorname{Construct}(\mathsf{DB},\mathsf{Docs},\mathsf{Examples}).
\label{eq:construct}
\end{equation}
Here, \(\operatorname{Construct}\) denotes the ontology construction process, and \(O_0\) denotes the resulting initial ontology, with the subscript \(0\) distinguishing it from subsequent ontology states.

Following Equation~\ref{eq:ontology}, $O_0$ consists of $E_0$, $M_0$, and $G_0$, which respectively denote its entities and relationships, metric definitions, and database mappings. AgentOnto constructs these components together by identifying business entities and relationships, organizing reusable metric definitions, and establishing database mappings that connect them to the corresponding tables and columns in $\mathsf{DB}$. The construction process aligns descriptions of the same enterprise knowledge across sources.
For example, a business concept described in $\mathsf{Docs}$ can be connected to its analytical use in $\mathsf{Examples}$ and the corresponding tables and columns in $\mathsf{DB}$.

For each constructed ontology entry $x$, AgentOnto records its supporting sources $\mathcal{S}(x)$ and confirmation status $c(x)$. The sources connect the entry to specific evidence in $\mathsf{DB}$, $\mathsf{Docs}$, or $\mathsf{Examples}$. A separate model confirmation step evaluates the entry against this evidence and records the result in $c(x)$. It checks whether entities, relationships, and metric definitions are supported by the available resources and whether database mappings connect them to the corresponding data in $\mathsf{DB}$.

Constructing the three ontology components together connects enterprise knowledge and analysis patterns to the corresponding data in $\mathsf{DB}$. The resulting ontology can therefore directly guide subsequent analysis over enterprise data.

\paratitle{Ontology-Guided Analysis.}
Ontology-Guided Analysis applies the enterprise analytics ontology throughout the processing of an analytical request. Given request $\mathsf{req}_i$, the ontology $O_t$ available when the request is received, and enterprise database $\mathsf{DB}$, AgentOnto performs
\begin{equation}
(\mathsf{action}_i,\mathsf{plan}_i,\mathsf{result}_i)
=
\operatorname{Analyze}(\mathsf{req}_i,O_t,\mathsf{DB}).
\label{eq:analyze}
\end{equation}
Here, $\operatorname{Analyze}$ denotes the complete analysis process. The subscript $i$ identifies the request and its corresponding outputs, while the subscript $t$ identifies the ontology state used for the analysis. The output $\mathsf{action}_i$ records the decision to execute the request, request clarification, or report that the request is unsupported. For an executable request, $\mathsf{plan}_i$ denotes its analysis plan and $\mathsf{result}_i$ denotes the result produced from $\mathsf{DB}$.

AgentOnto forms $\mathsf{plan}_i$ by connecting the requirements in $\mathsf{req}_i$ to the relevant content in $O_t$. The plan identifies the relevant ontology content and records the analytical choices specific to the request. These choices are represented through references to the ontology, while $G_t$ provides the database mappings needed to connect them to tables, columns, and join paths in $\mathsf{DB}$. When necessary information is missing or the available ontology and data cannot support the request, AgentOnto uses the identified limitation to determine the corresponding analytical action.

For an executable request, AgentOnto checks the analysis plan against the three components of $O_t$:
\begin{equation}
\mathsf{valid}_i
=
\operatorname{Check}(\mathsf{plan}_i;E_t,M_t,G_t).
\label{eq:check}
\end{equation}
Here, $\operatorname{Check}$ denotes the plan-checking process, and $\mathsf{valid}_i$ indicates whether the plan is consistent with the ontology. The check verifies that the plan uses defined entities and relationships from $E_t$, follows the metric requirements in $M_t$, and has the database mappings in $G_t$ required for execution. This step identifies analytical errors that successful database execution alone may not reveal.

A valid plan is compiled and executed as
\begin{equation}
\begin{aligned}
\mathsf{query}_i
&=
\operatorname{Compile}(\mathsf{plan}_i;G_t),\\
\mathsf{result}_i
&=
\operatorname{Execute}(\mathsf{query}_i;\mathsf{DB}).
\end{aligned}
\label{eq:compile-execute}
\end{equation}
The operator $\operatorname{Compile}$ uses the database mappings in $G_t$ to translate $\mathsf{plan}_i$ into the database query $\mathsf{query}_i$. The operator $\operatorname{Execute}$ runs this query over $\mathsf{DB}$ to produce $\mathsf{result}_i$. Only a plan that passes the ontology-based check proceeds to compilation and execution.

Through this process, the same enterprise analytics ontology guides plan formation, plan checking, and database execution. The explicit analysis plan connects the interpretation of a request to the query executed over enterprise data, allowing ontology knowledge to directly affect both analytical decisions and results.

\paratitle{Ontology Updating.}
Ontology Updating incorporates confirmed analytical feedback into the enterprise analytics ontology. Given the current ontology $O_t$, confirmed feedback $\mathsf{Feedback}_t$, update checks $\mathsf{Checks}_t$, and enterprise database $\mathsf{DB}$, AgentOnto performs
\begin{equation}
O_{t+1}
=
\operatorname{Update}
(O_t,\mathsf{Feedback}_t,\mathsf{Checks}_t,\mathsf{DB}).
\label{eq:update}
\end{equation}
Here, $\operatorname{Update}$ denotes the ontology updating process. The input $\mathsf{Feedback}_t$ contains the confirmed feedback available at state $t$, and $\mathsf{Checks}_t$ contains the analytical checks used to evaluate its effect. The output $O_{t+1}$ denotes the ontology after updating, with the subscript $t+1$ identifying the state following $O_t$.

Each item in $\mathsf{Feedback}_t$ identifies the ontology entry associated with the observed problem, the required change, and the evidence confirming that change. The target may belong to the entities and relationships $E_t$, metric definitions $M_t$, or database mappings $G_t$. AgentOnto uses this information to make a targeted modification to the relevant ontology content.

The update checks contain two complementary groups:
\begin{equation}
\mathsf{Checks}_t
=
\mathsf{Checks}_t^{\mathrm{corr}}
\cup
\mathsf{Checks}_t^{\mathrm{ret}}.
\label{eq:update-checks}
\end{equation}
The superscript $\mathrm{corr}$ denotes correction checks, which specify the analytical behavior that should change after the update. The superscript $\mathrm{ret}$ denotes retention checks, which specify the analytical behavior that should remain unchanged. Each check contains an analytical request and its expected outcome, including the expected analytical decision and, for an executable request, the expected result.

AgentOnto applies the targeted modification to produce a proposed ontology $\widetilde{O}_{t+1}$ and evaluates both groups of checks through Ontology-Guided Analysis. The tilde distinguishes this proposed ontology from the ontology adopted after checking:
\begin{equation}
O_{t+1}
=
\begin{cases}
\widetilde{O}_{t+1},
& \text{if }
\operatorname{CheckUpdate}
(\widetilde{O}_{t+1},\mathsf{Checks}_t,\mathsf{DB})
=
\mathrm{true},\\
O_t,
& \text{otherwise}.
\end{cases}
\label{eq:publish}
\end{equation}
The operator $\operatorname{CheckUpdate}$ evaluates the proposed ontology using the correction and retention checks. AgentOnto adopts $\widetilde{O}_{t+1}$ when it corrects the targeted problem and preserves the analytical behavior that should remain unchanged. Otherwise, it retains $O_t$.

Ontology Updating therefore evaluates a change through its effect on subsequent analysis. Confirmed feedback produces a targeted modification to reusable enterprise knowledge, while correction and retention checks ensure that the modification has its intended analytical effect.

\subsection{Iterative Ontology Preparation}

Ontology Construction integrates the available enterprise resources into an initial ontology. This initial ontology may still contain missing or incorrect knowledge that affects subsequent analysis. Iterative Ontology Preparation examines and refines the ontology through self-generated analysis tasks before actual analysis tasks.

Starting from $O_0$, AgentOnto performs $R$ preparation rounds and generates $B$ analysis tasks in each round. Here, $R$ denotes the total number of rounds and $B$ denotes the number of tasks generated per round. In round $r$, AgentOnto generates the task set $\mathsf{PrepTasks}_r$ from $\mathsf{DB}$, $\mathsf{Docs}$, $\mathsf{Examples}$, and the ontology $O_{r-1}$ produced by the preceding round. These tasks examine whether the current ontology can support analysis across its entities, relationships, metrics, and database mappings.

AgentOnto analyzes the generated tasks using $O_{r-1}$. A separate model confirmation step reviews the analytical outcomes against the enterprise resources and produces confirmed feedback $\mathsf{Feedback}_r$ for missing or incorrect ontology content. AgentOnto then derives correction and retention checks $\mathsf{Checks}_r$ and updates the ontology:
\begin{equation}
O_r
=
\operatorname{Update}
(O_{r-1},\mathsf{Feedback}_r,\mathsf{Checks}_r,\mathsf{DB}),
\qquad r=1,\ldots,R.
\label{eq:preparation}
\end{equation}
Here, $O_r$ denotes the ontology produced in round $r$. The update corrects the confirmed problem while checking the analytical behavior that should remain unchanged.
The updated ontology $O_r$ is used to generate and analyze tasks in the next round. After $R$ rounds, the resulting ontology $O_R$ has been repeatedly examined and refined through analytical use and is used for subsequent analysis tasks.

Iterative Ontology Preparation further examines whether the constructed ontology can support data analysis. Self-generated tasks expose problems that arise when ontology knowledge is applied and combined during analysis, even before feedback from actual tasks is available. Confirmed problems are incorporated into the reusable ontology through checked updates, allowing the resulting corrections to support subsequent requests.

\subsection{AgentOnto Workflow}

AgentOnto first constructs and prepares the enterprise analytics ontology, then uses the resulting ontology for actual analysis and subsequent updating.

The workflow connects the three core methods through the enterprise analytics ontology. Preparation examines and refines the initially constructed ontology, while subsequent updating incorporates confirmed feedback so that later requests use the corrected enterprise knowledge.

%% file: sections/4_exp.tex
\section{Experiments} \label{sec:exp}

We evaluate the complete AgentOnto workflow and its three core methods through four research questions:

\begin{itemize}[leftmargin=*, topsep=0pt]
\item \textbf{RQ1 (Overall Effectiveness):} How effectively does the complete AgentOnto workflow handle enterprise analysis requests?
\item \textbf{RQ2 (Ontology Construction):} To what extent does Ontology Construction organize heterogeneous enterprise knowledge into an ontology that connects to data and supports analysis?
\item \textbf{RQ3 (Ontology-Guided Analysis):} Given a fixed ontology, how does Ontology-Guided Analysis affect the correctness of analytical decisions and execution?
\item \textbf{RQ4 (Ontology Updating):} Can Ontology Updating correct a targeted problem while preserving analytical behavior that should remain unchanged?
\end{itemize}

\subsection{Experimental Setup}

\paratitle{Datasets.}
We evaluate AgentOnto on two public data sources from different domains. The World Development Indicators (WDI)~\cite{worldbank_wdi} contains country- and region-level development data, such as population and economic indicators across years. The U.S. EPA Air Quality System (AQS)~\cite{epa_aqs} contains measurements of pollutants such as ozone and PM2.5 collected at monitoring sites over time. These two domains provide different enterprise knowledge, data relationships, and analytical requests for evaluating AgentOnto.

For each domain, we prepare an enterprise database, business documents, and example analysis tasks that show how enterprise knowledge is applied to data. The benchmark contains 606 analysis requests, including 282 for WDI and 324 for AQS. It also includes 44 confirmed-feedback events and 174 post-update analysis requests for evaluating ontology updating.

\paratitle{Baseline Selection.}
The four research questions evaluate different capabilities, so we select baselines separately for each RQ. We use the released implementations and replace their invoked LLM with the model used by AgentOnto. The specific baselines are introduced in the corresponding RQ subsections.

\paratitle{Evaluation Protocol.}
We evaluate each method for ontology content, database connections, analytical decisions and results, or update effects, depending on the RQ. The specific evaluation criteria and metrics are introduced in the corresponding RQ subsections. We report WDI and AQS separately and include their domain average as a summary. 

\paratitle{Implementation Details.}
We use the \texttt{gpt-5.6-terra} model for all generative methods. Separate \texttt{gpt-5.6-sol} calls are used for reference construction and independent review. AgentOnto performs two rounds of Iterative Ontology Preparation and generates four analysis tasks in each round.

\subsection{Overall Effectiveness (RQ1)}

We evaluate the complete AgentOnto workflow on all 606 evaluation requests after Ontology Construction and two rounds of Iterative Ontology Preparation.

\paratitle{Compared Methods.}
We compare AgentOnto with DIN-SQL~\cite{pourreza2023din}, ReFoRCE-v8~\cite{deng2025reforce}, MAC-SQL~\cite{wang2025mac}, and Vanna~\cite{vanna_software}. Each comparison method performs its own preparation before processing the same requests. DIN-SQL, ReFoRCE-v8, and MAC-SQL represent multi-stage SQL generation methods, while Vanna represents database-agent tool use.

\paratitle{Metric.}
We use analysis accuracy to evaluate the complete workflow. A request is correct when the analytical decision is correct and, if execution is required, the returned result is also correct.

\paratitle{Results.}
AgentOnto correctly handles 223/282 WDI requests and 311/324 AQS requests, achieving analysis accuracy of 79.1\% and 96.0\%. These results exceed the strongest baseline in each domain by 32.6 and 30.3 percentage points, respectively. Its domain average reaches 87.5\%, compared with 55.0\% for DIN-SQL. The consistent improvement across both domains shows that the complete AgentOnto workflow remains effective across different enterprise knowledge and data-analysis requirements. The following RQs examine how its three core methods contribute to this overall result.

\input{tables/rq1_end_to_end.tex}

\subsection{Ontology Construction (RQ2)}

We evaluate the enterprise analytics ontology constructed by each method from the same database, business documents, and example analysis tasks.

\paratitle{Compared Methods.}
We compare AgentOnto Construction with Single-Prompt LLM, OntoEKG~\cite{oyewale2026llm}, Memoryless CQ-by-CQ~\cite{lippolis2025ontology}, and Ontop~\cite{calvanese2016ontop}. We distinguish ontology content, which includes entities, relationships, and metric definitions, from database mappings, which connect this knowledge to tables and columns. AgentOnto Construction and Single-Prompt LLM construct both parts. OntoEKG and Memoryless CQ-by-CQ construct ontology content without database mappings, while Ontop constructs database mappings from the database schema.

\paratitle{Metrics.}
We use ontology-content F1, database-connection accuracy, and analysis accuracy. Ontology-content F1 measures the correctness and coverage of the constructed ontology content. Database-connection accuracy measures whether its mappings reach the required data. Analysis accuracy measures whether the constructed ontology supports correct analytical decisions and results when used by the same analysis method. Metrics for capabilities not supported by a method are marked as ``--''.

\input{tables/rq2_construction.tex}

\paratitle{Results.}
AgentOnto achieves ontology-content F1 of 93.6\% on WDI and 76.7\% on AQS. It passes 43/51 WDI and 29/42 AQS database-connection tests, corresponding to 84.3\% and 69.0\%. Compared with Single-Prompt LLM, AgentOnto improves content F1 by 10.6 and 17.6 points and database-connection accuracy by 41.2 and 21.4 percentage points across the two domains. The improvements in analysis accuracy are smaller at 2.1 points on WDI and 1.9 points on AQS, showing that the largest gains occur in the constructed ontology and its connection to data.

Methods that construct only one part of the required ontology show limitations. OntoEKG and Memoryless CQ-by-CQ do not produce database mappings and achieve analysis accuracy between 16.4\% and 22.0\%. Ontop reaches 88.1\% database-connection accuracy on AQS, but its analysis accuracy is 14.5\% without the enterprise knowledge needed to interpret requests. These results show that ontology content and database mappings both contribute to an ontology that can support enterprise data analysis.
\input{tables/rq3_analysis.tex}

\subsection{Ontology-Guided Analysis (RQ3)}

We evaluate Ontology-Guided Analysis on all 606 evaluation requests. WDI contains 206 executable requests, 30 requests requiring clarification, and 46 unsupported requests. AQS contains 250 executable requests, 28 requests requiring clarification, and 46 unsupported requests.

\paratitle{Compared Methods.}
We compare AgentOnto Analysis with Direct Schema, Direct Documents, Direct Ontology, DIN-SQL~\cite{pourreza2023din}, ReFoRCE-v8~\cite{deng2025reforce}, MAC-SQL~\cite{wang2025mac}, and Vanna~\cite{vanna_software}. The three direct methods generate analytical decisions and queries directly from their corresponding context. AgentOnto Analysis receives the same inputs as Direct Ontology and uses the ontology throughout the analytical process. DIN-SQL, ReFoRCE-v8, and MAC-SQL represent multi-stage SQL generation methods, while Vanna represents a database-agent method.

\paratitle{Metrics.}
We use analysis accuracy and silent semantic error rate as the main metrics. Analysis accuracy requires the correct analytical decision and, for an executable request, the correct result. Silent semantic error rate measures how often a method returns an executable result with an incorrect decision or result. 
We also report answer coverage, the proportion of requests receiving executable results. Higher coverage does not necessarily indicate better analysis because not every request should be executed.

\paratitle{Results.}
AgentOnto Analysis achieves analysis accuracy of 93.6\% on WDI and 98.1\% on AQS, exceeding the strongest comparison method in each domain by 47.1 and 32.4 percentage points. Its domain average reaches 95.9\%, compared with 55.0\% for DIN-SQL. The improvement in both domains demonstrates the effectiveness of Ontology-Guided Analysis across different analytical requirements.

Using the same ontology and database inputs, Direct Ontology achieves domain-average analysis accuracy of 51.4\% and a silent semantic error rate of 50.9\%. AgentOnto Analysis achieves 95.9\% and 5.5\% on the same metrics. This comparison shows that applying the ontology throughout the analytical process provides more reliable analysis than using it only as model context.

AgentOnto Analysis has domain-average answer coverage of 75.7\%, which is close to the proportion of requests that should be executed. The multi-stage SQL and database-agent methods return executable results for more than 92\% of requests, but their silent semantic error rates range from 44.0\% to 56.5\%. These results show that answer coverage alone does not indicate correct analysis.

\subsection{Ontology Updating (RQ4)}

We evaluate Ontology Updating on 22 confirmed-feedback events from AQS. The evaluation includes 29 correction, 29 behavior-preservation, and 29 generalization requests.

\paratitle{Compared Methods.}
Static retains the existing ontology, Context-Only provides feedback with subsequent requests, and Memory/\allowbreak Reflection uses an LLM to interpret the feedback and stores the resulting reflection as textual memory for subsequent requests.

\paratitle{Metrics.}
We evaluate the updated ontology and its subsequent analysis against fixed reference records. Update accuracy compares the resulting ontology with the expected update. Correction and generalization accuracy measure whether the target problem is corrected and the updated knowledge supports related requests. Behavior Preservation is the proportion of requests that remain correct after the update among those answered correctly before it. Update accuracy is shown as ``--'' for methods that do not produce an updated ontology.

\input{tables/rq4_updating.tex}

\paratitle{Results.}
AgentOnto Updating correctly handles all 22 feedback events and all 29 requests in each behavioral category, achieving 100.0\% on all four metrics. Context-Only and Memory/\allowbreak Reflection reach 89.7\% correction and generalization accuracy, but their Behavior Preservation is 89.3\% and 86.2\%. Static cannot incorporate the feedback and achieves only 3.4\% correction and 17.2\% generalization accuracy. Its ontology remains unchanged, yet its Behavior Preservation is 22/29 (75.9\%) because repeated LLM analysis fails to reproduce seven results that were correct before the event. This metric therefore measures preservation of analytical behavior across repeated analysis. AgentOnto is the only evaluated method that consistently combines correct ontology updates, correction, behavior preservation, and generalization.

%% file: tables/rq1_end_to_end.tex
\begin{table}[t]
  \centering
  \caption{Analysis accuracy on WDI and AQS and the domain average. Bold and underlined values denote the best and second-best results.}
  \label{tab:rq1-end-to-end}
  \small
  \resizebox{\linewidth}{!}{
  \renewcommand{\arraystretch}{1.01}
  \begin{tabular}{l|cc|cc|c}
    \toprule
    \multirow{2}{*}{Method}
    & \multicolumn{2}{c|}{WDI ($N=282$)}
    & \multicolumn{2}{c|}{AQS ($N=324$)}
    & \multirow{2}{*}{\makecell{Domain \\ Average (\%)}} \\
    \cmidrule(lr){2-3} \cmidrule(lr){4-5}
    & Correct & Accuracy (\%) & Correct & Accuracy (\%) & \\
    \midrule
    DIN-SQL & 125 & 44.3 & \underline{213} & \underline{65.7} & \underline{55.0} \\
    ReFoRCE-v8 & 122 & 43.3 & 205 & 63.3 & 53.3 \\
    MAC-SQL & 125 & 44.3 & 142 & 43.8 & 44.1 \\
    Vanna & \underline{131} & \underline{46.5} & 131 & 40.4 & 43.4 \\
    \midrule
    \textbf{AgentOnto} & \textbf{223} & \textbf{79.1} & \textbf{311} & \textbf{96.0} & \textbf{87.5} \\
    \bottomrule
  \end{tabular}
  }
\end{table}

%% file: tables/rq2_construction.tex
\begin{table*}[t]
  \centering
  \caption{Ontology construction results on WDI and AQS, grouped by method capability. A - denotes an unsupported capability. Bold and underlined values denote the best and second-best results in each column.}
  \resizebox{\textwidth}{!}{
    \renewcommand{\arraystretch}{1.01}
    \begin{tabular}{ll|ccc|ccc}
      \toprule
      \multirow{2}{*}{Capability} & \multirow{2}{*}{Method}
      & \multicolumn{3}{c|}{WDI}
      & \multicolumn{3}{c}{AQS} \\
      \cmidrule(lr){3-5} \cmidrule(lr){6-8}
      & & \makecell{Ontology\\F1 (\%)}
        & \makecell{Database\\connection (\%)}
        & \makecell{Analysis\\accuracy (\%)}
      & \makecell{Ontology\\F1 (\%)}
        & \makecell{Database\\connection (\%)}
        & \makecell{Analysis\\accuracy (\%)} \\
      \midrule
      Ontology content + mappings
        & \textbf{AgentOnto Construction}
        & \textbf{93.6} & \textbf{84.3} & \textbf{70.2}
        & \textbf{76.7} & \underline{69.0} & \textbf{67.0} \\
      Ontology content + mappings
        & Single-Prompt LLM
        & \underline{83.0} & 43.1 & \underline{68.1}
        & \underline{59.1} & 47.6 & \underline{65.1} \\
        \midrule
      Ontology content only
        & OntoEKG
        & 2.9 & - & 22.0
        & 5.3 & - & 16.4 \\
      Ontology content only
        & Memoryless CQ-by-CQ
        & 6.5 & - & 20.6
        & 25.7 & - & 16.7 \\
      \midrule
      Database mappings only
        & Ontop bootstrap
        & - & \underline{51.0} & 15.6
        & - & \textbf{88.1} & 14.5 \\
      \bottomrule
    \end{tabular}}
\end{table*}

%% file: tables/rq3_analysis.tex
  \begin{table*}[t]
    \centering
  \caption{Analysis results on WDI and AQS (\%) and their domain average. Higher analysis accuracy and lower silent error indicate better performance. Answer coverage provides additional context because not every request should be executed. Bold and underlined values denote the best and second-best results for analysis accuracy and silent error.}
    \label{tab:rq3-analysis}
    \resizebox{1\linewidth}{!}{
      \renewcommand{\arraystretch}{1}
      \begin{tabular}{l|ccc|ccc|ccc}
        \toprule
        \multirow{2}{*}{Method}
        & \multicolumn{3}{c|}{WDI}
        & \multicolumn{3}{c|}{AQS}
        & \multicolumn{3}{c}{Domain Average} \\
        \cmidrule(lr){2-4} \cmidrule(lr){5-7} \cmidrule(lr){8-10}
        & \makecell{Analysis\\accuracy $\uparrow$}
        & \makecell{Silent\\error $\downarrow$}
        & \makecell{Answer\\coverage}
        & \makecell{Analysis\\accuracy $\uparrow$}
        & \makecell{Silent\\error $\downarrow$}
        & \makecell{Answer\\coverage}
        & \makecell{Analysis\\accuracy $\uparrow$}
        & \makecell{Silent\\error $\downarrow$}
        & \makecell{Answer\\coverage} \\
        \midrule
        Direct Schema & 37.9 & 63.0 & 74.8 & 8.6 & 97.3 & 78.7 & 23.3 & 80.1 & 76.8 \\
        Direct Documents & 45.4 & 60.2 & 83.7 & 34.9 & 74.4 & 84.3 & 40.1 & 67.3 & 84.0 \\
        Direct Ontology & 43.6 & 60.2 & 78.4 & 59.3 & 41.6 & 84.6 & 51.4 & 50.9 & 81.5 \\
        DIN-SQL & 44.3 & 55.2 & 98.9 & \underline{65.7} & \underline{32.8} & 97.8 & \underline{55.0} & \underline{44.0} & 98.4 \\
        ReFoRCE-v8 & 43.3 & 56.7 & 100.0 & 63.3 & 36.3 & 99.4 & 53.3 & 46.5 & 99.7 \\
        MAC-SQL & 44.3 & 54.5 & 97.5 & 43.8 & 49.3 & 86.4 & 44.1 & 51.9 & 92.0 \\
        Vanna & \underline{46.5} & \underline{53.5} & 100.0 & 40.4 & 59.4 & 99.7 & 43.4 & 56.5 & 99.8 \\
        \midrule
        \textbf{AgentOnto Analysis}
        & \textbf{93.6} & \textbf{8.7} & 73.4
        & \textbf{98.1} & \textbf{2.4} & 78.1
        & \textbf{95.9} & \textbf{5.5} & 75.7 \\
        \bottomrule
      \end{tabular}}
  \end{table*}

%% file: tables/rq4_updating.tex
  \begin{table}[t]
    \centering
    \caption{Ontology updating results on 22 AQS feedback events and 29 requests per behavioral category (\%). Higher values indicate better performance. A dash indicates that a method does not produce an updated ontology.}
    \label{tab:rq4-updating}
    \resizebox{\columnwidth}{!}{
      \renewcommand{\arraystretch}{1.05}
      \begin{tabular}{l|cccc}
        \toprule
        Method & Update & Correction & \makecell{Behavior\\Preservation} & Generalization \\
        \midrule
        Static
        & 22.7 & 3.4 & 75.9 & 17.2 \\
        Context-Only
        & -- & 89.7 & 89.3 & 89.7 \\
        Memory/Reflection
        & -- & 89.7 & 86.2 & 89.7 \\
        \midrule
        \textbf{AgentOnto Updating}
        & \textbf{100.0}
        & \textbf{100.0}
        & \textbf{100.0}
        & \textbf{100.0} \\
        \bottomrule
      \end{tabular}}
  \end{table}

%% file: sections/5_deployment.tex
\section{Enterprise Deployment} \label{sec:deploy}

AgentOnto has been deployed for enterprise analytics in consumer manufacturing, higher education, and transportation research.
We report post-launch results from the consumer-manufacturing deployment, which supports market-intelligence and device-data analytics.
The market-intelligence data cover 10 product categories, with the longest fact table spanning 53 consecutive months.
The device data contain approximately 437M reports from 96 device models and are aggregated into 333 ontology objects.
The higher-education deployment covers 3,000 students and 36,000 grade records, while the transportation-research deployment contains approximately 53.5K aggregate records across seven business object types.

\begin{figure}[t]
    \centering
    \begin{subfigure}[t]{0.52\columnwidth}
        \centering
        \includegraphics[width=\linewidth]{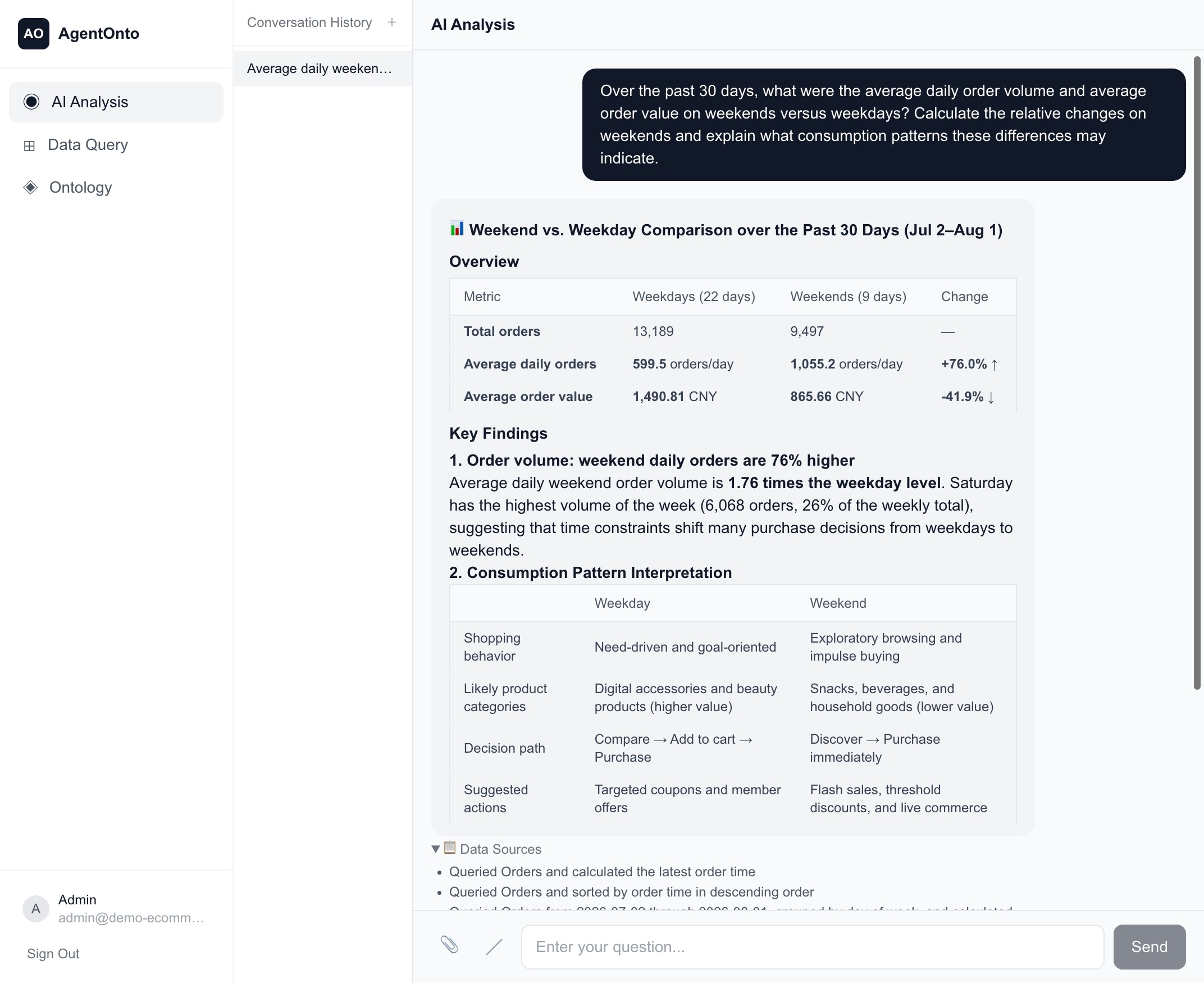}
        \caption{Deployed analytics interface}
        \label{fig:deploy-interface}
    \end{subfigure}%
    \hfill
    \begin{subfigure}[t]{0.45\columnwidth}
        \centering
        \includegraphics[width=\linewidth]{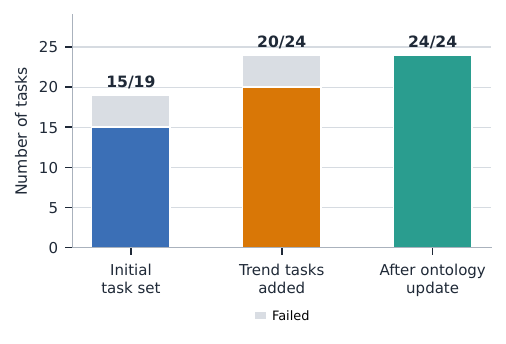}
        \caption{Controlled task evaluation}
        \label{fig:deploy-evaluation}
    \end{subfigure}
    \caption{Enterprise deployment in consumer manufacturing. (a) Deployed interface presenting an analysis of order volume, order value, and consumption patterns. (b) Controlled task results with task expansion and ontology updating.}
    \label{fig:deployment-postlaunch}
\end{figure}

The consumer-manufacturing deployment enabled AgentOnto's Ontology Construction, Ontology-Guided Analysis, and Ontology Updating.
Figure~\ref{fig:deploy-interface} shows the interface, where AgentOnto analyzes order volume and value and explains consumption patterns.
During a one-week observation window, 15 customer accounts completed 144 analytical sessions, in which the system performed 918 data operations. Aggregation result coverage, measured as the proportion of aggregation operations returning a non-empty result, increased from 453/541 (84\%) in pre-deployment testing to 106/106 (100\%) after deployment with real users. End-to-end latency was typically 30--60 seconds, with a 90th percentile of 46 seconds. AgentOnto therefore supported multi-user, cross-session analytics.
We further conducted a controlled task evaluation across three stages. AgentOnto passed 15/19 tasks on the initial set and 20/24 after trend-analysis tasks were added. Following a confirmed ontology update, it passed all 24 tasks. The update added a missing rule for calculating total share across price segments and corrected failures in four scenarios. These results show that ontology rules guide deployed analysis and that targeted updates correct the affected analytical behavior.
The consumer manufacturer subsequently integrated a second product-line domain within the same day. This integration added its data and analytical definitions without changing AgentOnto's core methods, showing how ontology extension supports new domains through the same analysis process.

%% file: sections/6_related.tex
\section{Related Work} \label{sec:related}

\subsection{LLM Agents for Data Analytics}

LLM-based data analytics has progressed from prompt-based text-to-SQL toward agents that retrieve context, decompose requests, use database tools, and incorporate execution feedback. Existing methods improve request decomposition, example selection, schema exploration, multi-agent coordination, and execution-guided correction~\cite{pourreza2023din,gao2023text,wang2025mac,deng2025reforce}. Enterprise BI systems further connect these capabilities with enterprise knowledge. SiriusBI~\cite{jiang2024siriusbi} retrieves knowledge and produces an intermediate representation, while CoeusBI~\cite{lian2026coeusbi} constructs reusable views and compiles a unified query representation. These systems demonstrate the value of context retrieval, tool use, and reusable analytical knowledge. Enterprise knowledge, however, remains distributed across context, prompts, views, and memory. AgentOnto organizes this knowledge as an explicit ontology, uses it throughout analysis, and incorporates confirmed feedback through updates.

\subsection{Ontology-Based Data Management}

Ontology-based data management organizes domain knowledge and connects it to databases. Ontop~\cite{calvanese2016ontop} and the OBDA literature~\cite{xiao2018ontology} combine ontologies, database mappings, and query rewriting. Existing systems make ontology-based access available through terms and natural-language questions~\cite{blunschi2012soda,saha2016athena,sen2020athena++,kharlamov2017ontology}. Analytical semantic layers similarly organize entities, measures, and database relationships for query compilation~\cite{hyde2024measures,kim2026semantic,rumiantsau2026semantic}. AI-assisted ontology engineering uses language models and retrieval to identify concepts, relationships, definitions, and ontology structures from documents and existing knowledge. Representative systems include LLMs4OL~\cite{giglou2023llms4ol}, DRAGON-AI~\cite{toro2024dynamic}, OntoChat~\cite{zhang2024ontochat}, and methods for generating taxonomies or modular ontologies~\cite{funk2023towards,lo2024end,lippolis2025ontology,oyewale2026llm}. Towards Agentic Schema Refinement~\cite{rissaki2024towards} is closely related to Iterative Ontology Preparation because it generates analysis tasks and validates reusable database views through execution. Other studies apply ontologies to query answering and checking~\cite{sequeda2024benchmark,allemang2024increasing}, ontology matching~\cite{hertling2023olala}, and ontology evolution and change management~\cite{zablith2009evolva,tudorache2013webprotege,qiang2024om4ov,noy2004ontology,flouris2008ontology}. These capabilities are formulated as separate processes. AgentOnto integrates Ontology Construction, Ontology-Guided Analysis, and Ontology Updating through an enterprise analytics ontology. Iterative Ontology Preparation examines the ontology through self-generated analysis tasks before actual requests.

%% file: sections/7_conclusion.tex
\section{Conclusion}
\label{sec:conclusion}

We presented AgentOnto, an ontology-guided LLM agent that organizes heterogeneous enterprise knowledge into an enterprise analytics ontology, uses the ontology to guide analysis, and incorporates confirmed feedback through targeted updates. Iterative Ontology Preparation further improves the ontology through self-generated analysis tasks. 
Experiments on two public data sources show that AgentOnto improves overall analysis accuracy and the effectiveness of ontology construction, analysis, and updating, while a real-world deployment demonstrates its practical value.
The enterprise analytics ontology connects reusable enterprise knowledge to analytical decisions and allows confirmed improvements to support subsequent requests. These results establish ontology-guided LLM agents as a systematic approach to reliable and adaptable enterprise data analytics.